%% file: main.tex
\documentclass[letterpaper, 10 pt, conference]{IEEEtran} 
\IEEEoverridecommandlockouts

\usepackage{algorithm}
\usepackage[noend]{algpseudocode}
\usepackage{amsfonts}
\usepackage{amsmath}
\usepackage{amssymb}
\usepackage{booktabs}
\usepackage{comment}
\usepackage{dsfont}
\usepackage{flushend}
\usepackage{hyperref}
\usepackage{lipsum} 
\usepackage{mathtools}
\usepackage{multirow}
\usepackage[
    detect-all=true,
    detect-family=true,
    group-digits=false,
    separate-uncertainty=true
]{siunitx}
\usepackage{url}

\usepackage{vector} 
\usepackage{adjustbox}

\usepackage[
    style=ieee,
    backend=biber,
    url=false,
    doi=false,
    natbib=true,
    mincitenames=1,
    dashed=false,
    isbn=false
]{biblatex}
\usepackage[capitalize]{cleveref}

\AtEveryBibitem{\clearfield{issn}}
\AtEveryCitekey{\clearfield{issn}}

\usepackage{color}
\usepackage{graphicx}
\usepackage{pgfplots}
\pgfplotsset{compat=newest}
\pgfplotsset{every axis legend/.append style={legend cell align=left}}
\usepackage[mode=buildnew]{standalone}
\usepackage{tikz}
\usetikzlibrary{shapes, arrows, external, positioning}
\usepackage{wrapfig}

\definecolor{pastelBlue}{RGB}{0,114,178}
\definecolor{pastelRed}{RGB}{245,97,92}
\definecolor{pastelGreen}{RGB}{0,158,115}
\definecolor{pastelPurple}{RGB}{135,112,254}

\DeclarePairedDelimiterX{\infdivx}[2]{[}{]}{%
    #1\;\delimsize\|\;#2%
}

\renewcommand{\vec}[1]{\vect{#1}}
\newcommand{\mat}[1]{\vect{#1}}

\renewrobustcmd{\bfseries}{\fontseries{b}\selectfont}
\renewrobustcmd{\boldmath}{}
\newrobustcmd{\B}{\bfseries}

\newcommand{\branches}{\textsc{Branches}}
\newcommand{\duffing}{\textsc{Oscillator}}
\newcommand{\gcas}{\textsc{F-16 GCAS}}

\title{\LARGE \bf
Scalable Importance Sampling in High Dimensions with \\ Low-Rank Mixture Proposals
}

\author{Liam A. Kruse, Marc R. Schlichting, and Mykel J. Kochenderfer%
\thanks{L. A. Kruse, M. R. Schlichting, and M. J. Kochenderfer are with the Stanford Intelligent Systems Laboratory in the Department of Aeronautics and Astronautics at Stanford University, Stanford, CA 94305, USA (email: \{lkruse, mschl, mykel\}@stanford.edu). }
}

\begin{document}
\maketitle
\thispagestyle{empty}
\pagestyle{empty}

\input{s0-abstract}
\input{s1-introduction}
\input{s2-related-work}
\input{s3-preliminaries}
\input{s4-methodology}
\input{s5-experiments}
\input{s6-conclusion}

\section*{Acknowledgments}
Toyota Research Institute (TRI) provided funds to assist the authors with their research, but this article solely reflects the opinions and conclusions of its authors and not TRI or any other Toyota entity.

\renewcommand*{\bibfont}{\footnotesize}
\printbibliography

\end{document}

%% file: s0-abstract.tex
\begin{abstract} 
Importance sampling is a Monte Carlo technique for efficiently estimating the likelihood of rare events by biasing the sampling distribution towards the rare event of interest.
By drawing weighted samples from a learned \textit{proposal} distribution, importance sampling allows for more sample-efficient estimation of rare events or tails of distributions. 
A common choice of proposal density is a Gaussian mixture model (GMM).
However, estimating full-rank GMM covariance matrices in high dimensions is a challenging task due to numerical instabilities.
In this work, we propose using mixtures of probabilistic principal component analyzers (MPPCA) as the parametric proposal density for importance sampling methods.
MPPCA models are a type of low-rank mixture model that can be fit quickly using expectation-maximization, even in high-dimensional spaces.
We validate our method on three simulated systems, demonstrating consistent gains in sample efficiency and quality of failure distribution characterization.
\end{abstract}

%% file: s1-introduction.tex
\section{Introduction}
\label{sec:introduction}
\begin{figure}[ht]
    \centering
    \includegraphics[width=\columnwidth]{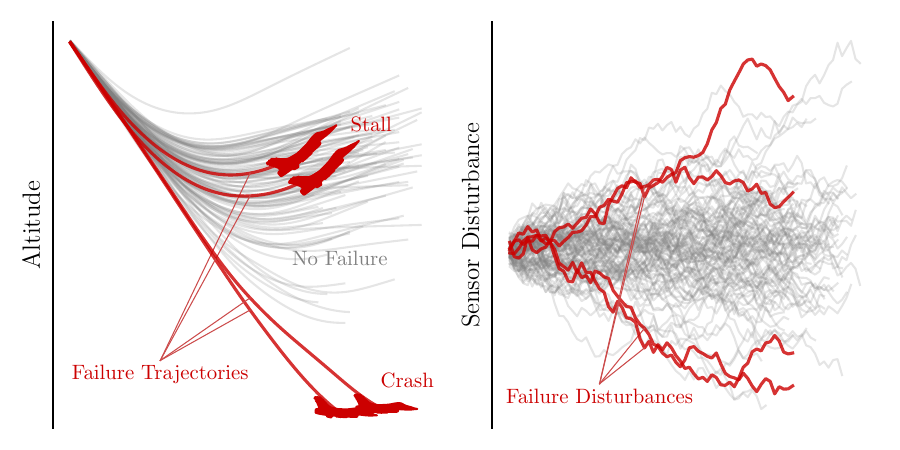}
    \caption{Safety-critical systems often have extremely low failure rates. Failures can arise from a sequence of disturbances, making the problem high-dimensional. While failures can be multimodal, the number of failure modes is typically much lower than the disturbance dimensionality. This motivates our approach of using low-rank mixture proposals to approximate the distribution over disturbances.}
    \label{fig:motivation}
\end{figure}

Safety-critical applications such as aircraft controller design or autonomous driving heavily rely on simulations to identify potential failures before real-world deployment.
Rigorous safety validation frameworks can characterize failure modes in a controlled simulation environment, reducing the risk of accidents \cite{corso2021survey}.
However, failure events such as collisions or loss of vehicle control might be rarely encountered in simulation due to strict safety thresholds, as shown in \cref{fig:motivation}.
\textit{Importance sampling} (IS) reduces the variance of Monte Carlo failure estimates by biasing the sampling distribution towards the rare event of interest \cite{mcbook, corso2021survey}.
IS uses a \textit{proposal distribution} to concentrate computational effort on scenarios likely to yield failure events, and assigns \textit{importance weights} to sampled points to correct for the biased distribution.
This efficient sampling technique reduces the variance of failure probability estimates compared to direct Monte Carlo sampling.

A common choice of parametric proposal density is a Gaussian mixture model (GMM) \cite{geyer2019cross}.
However, learning full-rank covariance matrices in high dimensions, especially when the number of dimensions is greater than the size of the dataset, is a challenging task.
The estimate might overfit to the noise in the dataset, and the matrices themselves become ill-conditioned or singular \cite{ledoit2004well}.
Furthermore, the memory required to store the matrices grows quadratically with the number of dimensions.

One solution is to constrain the learned covariance matrices to be low-rank, effectively modeling the covariances as full-rank matrices on a learned low-dimensional subspace \cite{tipping1999mixtures}.
In this work, we propose using mixtures of probabilistic principal component analyzers (MPPCA) as the parametric proposal density.
MPPCA models offer two key characteristics: 1) they have an analytical likelihood expression, preserving the computational efficiency required for computing importance weights, and 2) they can be fit efficiently even on high-dimensional data using the expectation-maximization (EM) algorithm. 
Our specific contributions include the following:
\begin{itemize}
    \item We construct expressive importance sampling proposal distributions using MPPCA models and demonstrate that the MPPCA fitting procedure is tractable even in high-dimensional domains. 
    \item We quantitatively evaluate our approach on simulated importance sampling tasks, including safety validation for an aircraft ground collision avoidance system.
\end{itemize}

%% file: s2-related-work.tex
\section{Related Work}
\label{sec:related-work}
Importance sampling has received widespread interest in applications such as structural reliability analysis \cite{papaioannou2016sequential, geyer2019cross}, safety validation for autonomous vehicles \cite{o2018scalable, huang2019evaluation}, and estimating the probability of failure for aircraft collision avoidance systems \cite{kim2016improving}.
Many IS methods such as the cross-entropy method or population Monte Carlo iteratively adapt the proposal distribution to move towards the failure distribution \cite{kochenderfer2024algorithms}.
The proposal distribution is often constrained to a parametric set of probability distributions such as the exponential family \cite{zhao2016accelerated, o2018scalable, kochenderfer2024algorithms}.
Parametric proposals admit analytical updates, which is computationally advantageous when refining the proposal to more closely match the failure distribution \cite{corso2021survey}.
The importance weights can be computed in closed form when the proposal density has an analytical likelihood expression.
However, iteratively refining proposals in high dimensions is challenging because the importance weights degenerate as the problem dimensionality increases \cite{botev2008efficient}.
Furthermore, full-rank covariance estimates can become numerically ill-conditioned in high dimensions, especially if the number of samples is small \cite{ledoit2004well}.
Neural network-based IS methods often scale more effectively to higher dimensions. 
\citet{demangevariational} approximate the failure distribution using a variational autoencoder, which requires iterative re-training throughout the IS procedure. 
Similarly, \citet{delecki2024diffusion} employ a denoising diffusion model as the proposal distribution, training it through a cross-entropy-like approach. 
While these methods produce high-fidelity results and perform well in high-dimensional settings, they are computationally expensive.
In this work, we use mixtures of probabilistic principal component analyzers as the parametric proposal density.
MPPCA models perform local linear dimensionality reduction, promoting scalable performance even in high dimensions.

Non-parametric approaches to IS offer flexible failure distribution representations that are not constrained to a specific distribution family. 
Sequential Monte Carlo methods can represent complex, multimodal failure distributions with a collection of samples \cite{del2006sequential, kochenderfer2024algorithms}.
Multilevel-splitting relies on Markov chain Monte Carlo (MCMC) estimation to guide a series of conditional distributions towards the failure distribution \cite{kahn1951estimation, norden2019efficient}.
However, MCMC can require many iterations to find all failure modes and accurately reflect the failure density \cite{kochenderfer2024algorithms}.
In general, non-parametric strategies do not admit closed-form updates.
We use MPPCA proposal densities because they can be analytically updated using the EM algorithm, resulting in a computationally efficient search over the space of sampling distributions.
Furthermore, we demonstrate that mixtures with even a small number of components are sufficiently expressive to model a range of failure modes.

Of particular interest to this work is the scalability of Gaussian mixture models.
Two common frameworks for modeling low-rank GMMs include mixtures of probabilistic principal component analyzers \cite{tipping1999mixtures} and mixtures of factor analyzers (MFAs) \cite{ghahramani1996algorithm}.
In this work, we focus on MPPCA models because they can be fit with closed-form expectation-maximization updates.
\citet{richardson2018gans} use MFA models for image generation, demonstrating that low-rank GMMs can be trained on full-sized images despite the high dimensionality.
Covariance matrix adaptation (CMA) is an evolutionary optimization algorithm that iteratively adapts a covariance matrix to improve sample efficiency \cite{hansen2016cma}.
Low-rank updates make the optimization process robust and sample efficient.

%% file: s3-preliminaries.tex
\section{Importance Sampling and MPPCA}
\label{sec:preliminaries}
We next outline the theory behind importance sampling and mixtures of probabilistic principal component analyzers.

\subsection{Importance Sampling}
\label{sec:is}
Simulations allow engineers to evaluate the performance of algorithms and models in diverse scenarios, including scenarios that are rare, dangerous, or costly to replicate in real-world testing.
Assessing the probability of failure events can require a prohibitively large number of Monte Carlo simulations, especially if the event of interest is rare.

Consider an outcome space $\mathbf{x} \in \mathbb{R}^d$ with probability density function $p(\mathbf{x})$ and a \textit{cost function} $f(\mathbf{x})$ such that a \textit{failure event} occurs if and only if $f(\mathbf{x}) \leq 0$.
The probability of failure $P_F$ is given by the integral
\begin{equation}
\label{eq:pf-integral}
    P_F = \mathbb{E}_{p(\mathbf{x})} \left[ \mathds{1} \{ f(\mathbf{x}) \leq 0 \} \right] = \int \mathds{1} \{ f(\mathbf{x}) \leq 0 \} \cdot p(\mathbf{x}) d\mathbf{x}
\end{equation}
We can estimate $P_F$ via Monte Carlo simulations by drawing $N_s$ samples $\lbrace \mathbf{x}_1, \dots, \mathbf{x}_{N_s} \rbrace$ from $p(\mathbf{x})$ and taking the mean:
\begin{equation}
\label{eq:pf-mcs-estimate}
    \hat{P}_F = \dfrac{1}{N_s} \sum_{n=1}^{N_s}  \mathds{1} \{ f(\mathbf{x}_n) \leq 0 \}.
\end{equation}
This estimate is unbiased and has a coefficient of variation
\begin{equation}
    \delta_{\hat{P}_F} = \sqrt{\dfrac{1 - P_F}{N_s P_F}}.
\end{equation} 
Since the coefficient of variation is inversely proportional to the failure probability, many samples might be required to precisely estimate $P_F$, especially if $P_F$ is small \cite{papaioannou2016sequential}.

Importance sampling is a technique that aims to reduce the variance of $\hat{P}_F$ by sampling from an alternative sampling distribution---or \textit{proposal distribution}---denoted by $q(\mathbf{x})$. 
So long as the support of $q(\mathbf{x})$ contains the failure domain, the probability of failure integral in \cref{eq:pf-integral} can be rewritten as
\begin{equation}
    P_F = \int \dfrac{\mathds{1} \{ f(\mathbf{x}) \leq 0 \} \cdot p(\mathbf{x}) }{q(\mathbf{x})} \cdot q(\mathbf{x}) d\mathbf{x}.
\end{equation}
The importance sampling estimate of $P_F$ is given by
\begin{equation}
\label{eq:is-pf}
    \hat{P}_F = \dfrac{1}{N_s} \sum_{n=1}^{N_s} \mathds{1} \{ f(\mathbf{x}_n) \leq 0 \} \cdot \dfrac{p(\mathbf{x}_n)}{q(\mathbf{x}_n)}
\end{equation}
where the samples are distributed according to the proposal distribution $q(\mathbf{x})$. 
An appropriate choice of proposal distribution can therefore reduce the variance of the estimate of $P_F$.

\subsection{Low-Rank Mixture Models}
\label{sec:mppca}
Principal component analysis (PCA) is a classic statistical technique for dimensionality reduction.
\citet{tipping1999mixtures} reformulate PCA within a maximum likelihood framework, resulting in an associated probability density.
Consider the following latent variable model:
\begin{align}
\label{eq:mppca}
\vec{x} &= \mat{W}\vec{z} + \vec{\mu} + \vec{\epsilon} \\
\vec{z} &\sim \mathcal{N}(\vec{0}, \mat{I}) \\
\vec{\epsilon} &\sim \mathcal{N}(\vec{0}, \sigma^2 \mat{I})
\end{align}
where $\mat{W}$ is a rectangular \textit{factor loading} matrix of size $d \times \ell$ and $\ell$ is a latent dimension such that $\ell \ll d$. 
The latent vector $\vec{z}$ is of length $\ell$, the mean vector $\vec{\mu}$ is of length $d$, and $\vec{\epsilon}$ is added noise with diagonal covariance $\sigma^2 \mat{I}$. 
For MPPCA, the noise is assumed to be isotropic; in the more general MFA framework, the noise assumption is relaxed to merely be diagonal.
In the case of isotropic noise, the implied conditional distribution is
\begin{equation}
    \label{eq:conditional}
    p\left(\vec{x} \mid \vec{z}\right) = \left( 2 \pi \sigma^2 \right)^{-d/2} \exp{\left\{ -\dfrac{1}{2\sigma^2} \lVert \vec{x} - \mat{W}\vec{z} - \vec{\mu}\rVert^2 \right\}}
\end{equation}
The Gaussian prior over the latent variables is given by
\begin{equation}
    \label{eq:prior}
    p\left(\vec{z}\right) = \left( 2 \pi \right)^{-\ell/2} \exp{\left\{ -\dfrac{1}{2} \vec{z}^\top \vec{z}\right\}}
\end{equation}
Thus, the marginal density over $\vec{x}$ can be expressed as
\begin{equation}
    \label{eq:marginal}
    p\left(\vec{x}\right) = \eta \lvert \mat{C} \rvert^{- 1 / 2} \exp{\left\{ -\dfrac{1}{2} \left(\vec{x} - \vec{\mu}\right)^\top \mat{C}^{-1} \left(\vec{x} - \vec{\mu}\right)\right\}}
\end{equation}
with normalizing constant $\eta = \left( 2 \pi \right)^{-d / 2}$ and model covariance $\mat{C} = \sigma^2 \mat{I} + \mat{W} \mat{W}^\top$.
\citet{tipping1999mixtures} show that the likelihood of a dataset $\{\vec{x}_0, \vec{x}_1, \ldots, \vec{x}_N\}$ is maximized when the columns of the factor loading matrix $\mat{W}$ span the principal subspace of the data.
\Cref{eq:marginal} defines the marginal likelihood for a single probabilistic principal component analyzer (PPCA); multiple PPCA models can be combined in a mixture to model more complex distributions.

%% file: s4-methodology.tex
\section{Methodology}
\label{sec:methodology}
In this section, we present the two adaptive importance sampling methods used in our study. 
We also describe the analytical EM procedure for optimizing proposal distributions.

\subsection{Importance Sampling Methods}
The cross-entropy (CE) method \cite{de2005tutorial, geyer2019cross} attempts to learn the parameters of a parametric proposal distribution that minimizes the KL divergence between the optimal IS density and the proposal.
CE importance sampling introduces a series of intermediate failure domains that gradually approach the true failure domain. 
At each step, the intermediate failure region is defined such that $\rho \cdot N_s$ samples fall in the region, where the $\rho$-quantile is chosen by the user.
The proposal distribution parameters are then fit via maximum likelihood estimation over these samples. 
The expectation-maximization algorithm is often used to fit the search distribution, though it must be adjusted to account for importance-weighted samples \cite{geyer2019cross}.

Like the CE method, sequential importance sampling (SIS) introduces a series of intermediate failure distributions that gradually approach the optimal IS density \cite{del2006sequential, papaioannou2016sequential}. 
Samples for each intermediate distribution are obtained by resampling weighted particles from the previous distribution and then moved to regions of high likelihood under the next failure distribution through Markov chain Monte Carlo. 
In this work, we use a conditional sampling Metropolis--Hastings algorithm to move the samples \cite{papaioannou2016sequential}.
The first samples in the Markov chains are discarded during a specified \textit{burn-in} period, ensuring that the retained samples are drawn from the stationary distribution.

\subsection{Fitting MPPCA Models}
Consider a mixture of $K$ probabilistic principal component analyzers as presented in \cref{eq:mppca}.
The log-likelihood of observing a dataset $\{\vec{x}_0, \vec{x}_1, \ldots, \vec{x}_N\}$ is
\begin{equation}
    \label{eq:loglike}
    \mathcal{L} = \sum_{i=1}^N \left\{ \sum_{k=1}^K \pi_k p\left(\vec{x}_n \mid k \right) \right\}
\end{equation}
where $p\left(\vec{x} \mid k \right)$ represents the density of the $k$th PPCA model and $\pi_k$ is the corresponding mixing proportion, with $\pi_k \geq 0$ and $\sum_{k=1}^K \pi_k = 1$.
\citet{tipping1999mixtures} derive an iterative expectation-maximization procedure with closed-form updates for MPPCA parameters $\mat{W}_k$, $\vec{\mu}_k$, $\pi_k$, and $\sigma^2_k$ for components $k = 1, \ldots, K$ that is guaranteed to find a local maximum of \cref{eq:loglike}.
In classical EM parameter updates for a mixture model, the \textit{responsibility} of component $k$ for generating sample $\vec{x}_n$ is given by 
\begin{equation}
r_{nk} = \dfrac{\pi_k p\left(\vec{x}_n \mid k\right)}{\sum_j \pi_k p\left(\vec{x}_n \mid j\right)}
\end{equation}
However, we must adjust the responsibility calculation since we are sampling from the proposal distribution \cite{geyer2019cross}.
The importance sampling responsibilities are
\begin{equation}
r_{nk} = \dfrac{w_n \pi_k q\left(\vec{x}_n \mid k\right)}{\sum_j \pi_k q\left(\vec{x}_n \mid j\right)}
\end{equation}
where $p(\mathbf{x}_n)$ is the prior density, $q(\mathbf{x}_n)$ is the proposal density, and $w_n = p(\mathbf{x}_n) / q(\mathbf{x}_n)$ is the \textit{likelihood ratio}.

%% file: s5-experiments.tex
\section{Experiments}
\label{sec:experiments}
This section presents our problem domains and evaluation metrics before discussing experimental results.

\subsection{Data Simulators}
We validate our proposed approach using three simulated environments. 
Each system is defined by a continuous cost function $f(\vec{x})$ that maps a $d$-dimensional sample $\vec{x}$ to a scalar value.
Recall from \cref{sec:mppca} that $f(\mathbf{x}) \leq 0$ defines a failure event, while positive values indicate that the sample falls outside of the failure region.
Larger values correspond to samples that are farther away from the failure region boundary.

\subsubsection{Branches}
\citet{chiron2023failure} develop the following analytical expression given for even dimensions with $d\geq2$: 
\begin{equation*}
    f\left(\mathbf{x}\right) = \min \left\{
    \begin{array}{c}
        \beta + \dfrac{1}{\sqrt{d}}\sum_{i=1}^d x_i\\
        \beta -\dfrac{1}{\sqrt{d}}\sum_{i=1}^d x_i\\
        \beta + \dfrac{1}{\sqrt{d}}\left(\sum_{i=1}^{d/2} x_i - \sum_{i=d/2+1}^{d} x_i\right)\\
        \beta + \dfrac{1}{\sqrt{d}}\left(-\sum_{i=1}^{d/2} x_i + \sum_{i=d/2+1}^{d} x_i\right)
    \end{array}
    \right\}
\end{equation*}
This system of four cost functions has a probability of failure which is independent of the number of random variables. 
We set the parameter $\beta=3.5$ to ensure that the probability of failure is small.

\subsubsection{Duffing Oscillator}
The second example is the Duffing oscillator introduced by \citet{zuev2009advanced}.
We consider its discretized form in the frequency domain as presented by \citet{papaioannou2019improved}. 
The cost function is the maximal displacement $u\left(t\right)$ of the oscillator at $t_{\text{max}} = 2$ seconds: 
\begin{equation*}
    f\left(\mathbf{x}\right) = \min\left\{u_1 - u\left(t_{\text{max}}\right), u\left(t_{\text{max}}\right)-u_2\right\}
\end{equation*}
We define $u_1 = 0.1$ and $u_2 = -0.06$, corresponding to two failure modes. 
The displacement of the oscillator satisfies 
\begin{align*}
    &m \ddot u\left(t\right) + c \dot u\left(t\right) + k\left(u\left(t\right) + \gamma u\left(t\right)^3\right) \\ &= -m\sigma\sum_{i=1}^{d/2} \left(x_i\cos\left(\omega_i t\right) + x_{d/2+i}\sin\left(\omega_i t\right)\right), 
\end{align*}
for all $t\geq 0$. We set $m =$ \SI{1000}{\kilogram}, $c = \SI[parse-numbers=false]{200\pi}{\newton\second/\meter}$, $k = \SI[parse-numbers=false]{1000(2\pi)^2}{\newton/\meter}$, $\gamma = \SI{1}{\per\meter\squared}$, $\omega_i = i\Delta\omega$, $\Delta\omega = 30\pi/d$, and $\sigma = \sqrt{0.01\Delta\omega}$.
The initial conditions are set to $u\left(0\right) = \SI{0}{\meter}$ and $\dot u\left(0\right) = \SI{1.5}{\meter /\second}$ \cite{demangevariational}.

\subsubsection{F-16 Ground Collision Avoidance}
Finally, we simulate a diving F-16 fighter jet controlled by a ground collision avoidance system (GCAS). 
We use the dynamics model introduced by \citet{heidlauf2018verification} and the \texttt{JAX} code implementation\footnote{\texttt{\url{https://github.com/MIT-REALM/jax-f16}}} by \citet{so2023solving}. 
In this experiment, we model the effect of a short-term sensor drift in the roll and pitch angle sensors. 
The F-16 GCAS system is activated at \SI{1000}{ft} and consists of two sequential phases: 1) leveling the wings and 2) increasing the pitch angle until the aircraft is recovered from the dive. 
The initial roll and pitch angles are sampled from $\phi_0\sim\mathcal{N}(0, 0.15^2)$ and $\theta_0\sim\mathcal{N}(-0.52, 0.05^2)$, respectively, while the initial altitude is \SI{950}{ft}. 
Roll and pitch sensor disturbances $\delta$ are modeled as a discrete-time approximation of the Wiener process $\delta_{t+1}\sim \delta_t+\epsilon$ with $\epsilon\sim\mathcal{N}(0,0.01^2)$. 
There are two possible failure modes: a collision with the ground (i.e., altitude $h\leq 0$) or an aerodynamic stall (i.e., angle of attack $\alpha\geq\alpha_c$). 
The critical angle of attack $\alpha_c$ is \SI{25}{degrees} for the F-16 model. 
The cost function for this experiment is
\begin{equation*}
    f(\mathbf{h}_{1:T}, \mathbf{\alpha}_{1:T})=\mathrm{minimum}\left(\dfrac{1}{950}\min_t\mathbf{h},\dfrac{1}{\alpha_c}\left(\alpha_c-\min_t \mathbf{\alpha}\right)\right)
\end{equation*}
The total dimensionality of the disturbance space is $d=202$.

\begin{figure*}[ht]
    \centering
    \includegraphics[width=\textwidth]{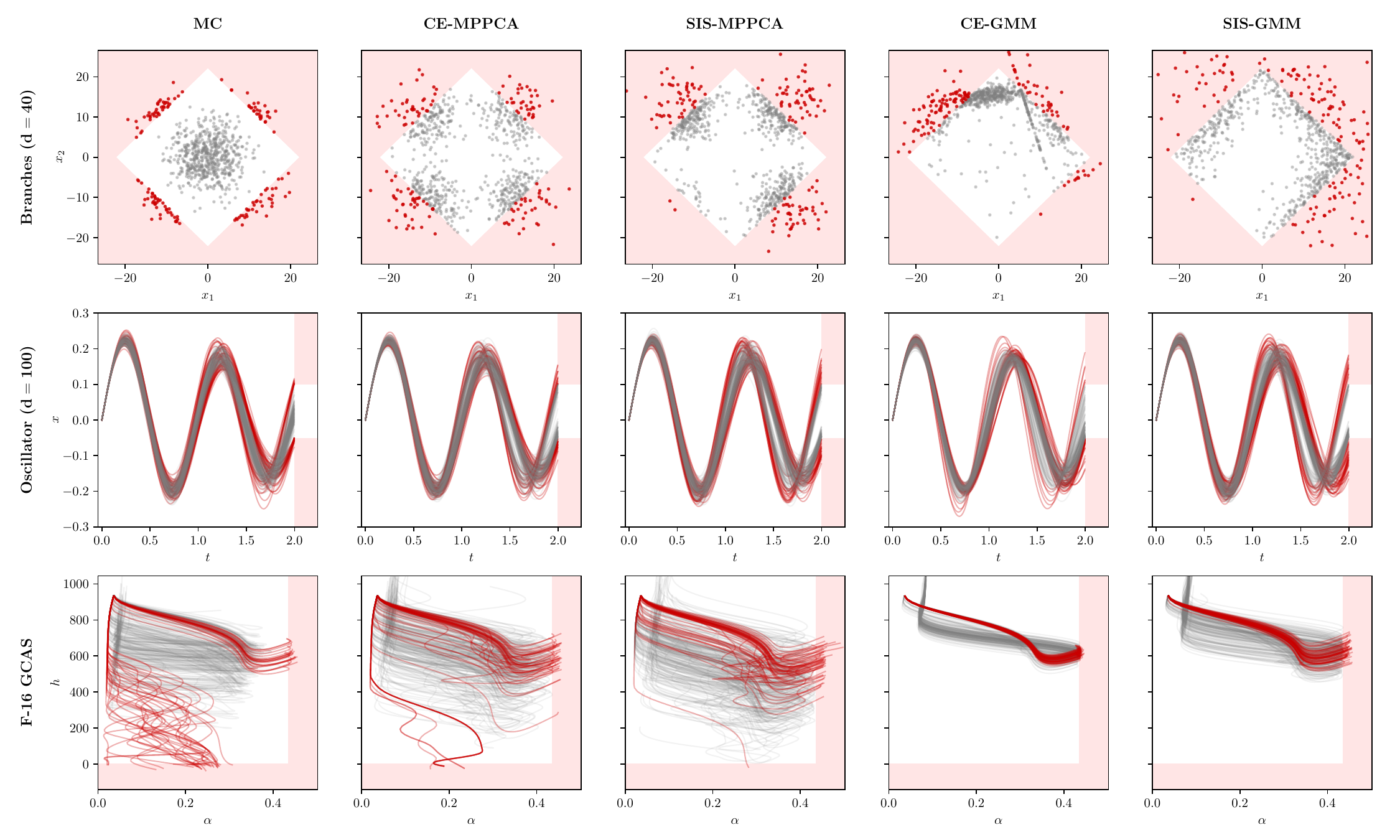}
    \caption{Results for the Branches problem with $d = 40$ (top row), the Duffing oscillator with $d = 100$ (middle row), and the F-16 GCAS system (bottom row). Red samples represent failure events where $f(\vec{x}) \leq 0$, while gray samples represent outcomes where $f(\vec{x}) > 0$. Across all experiments, the ratio of failure samples to non-failure samples is fixed at $1/4$, regardless of the system's failure rate or the effectiveness of the method. The shaded red areas denote the failure regions where the cost function is $\leq 0$.}
    \label{fig:results}
\end{figure*}

\begin{table}[!tb]
\centering
\caption{Hyperparameters}
\begin{tabular*}{\columnwidth}{@{\extracolsep{\fill}}lrrrrr@{}}
\toprule
System & $K$ & $L$ & samples / iter. & trials & $\rho$ \\ \midrule
Branches ($d=40$) & \num{8} & \num{8} & \num{10000} & \num{50} & \num{0.2} \\
Branches ($d=60$) & \num{8} & \num{8} & \num{10000} & \num{50} & \num{0.2} \\
Oscillator ($d=100$) & \num{8} & \num{8} & \num{10000} & \num{50} & \num{0.2} \\
Oscillator ($d=200$) & \num{8} & \num{8} & \num{10000} & \num{50} & \num{0.2} \\
F-16 GCAS & \num{8} & \num{8} & \num{10000} & \num{50} & \num{0.2} \\
\bottomrule
\end{tabular*}
\label{tab:hyperparameters}
\end{table}

\subsection{Metrics}
We obtain a reference failure probability for each dataset using the Monte Carlo estimate given in \cref{eq:pf-mcs-estimate} and then calculate the importance sampling estimate $\hat{P}_F$ using \cref{eq:is-pf}.
Next, we compute a series of metrics to evaluate the quality of the learned proposal density:
\begin{itemize}
    \item \textit{Relative error of $\hat{P}_F$}: This metric is defined as the signed difference between the IS estimate of the probability of failure and reference failure probability, normalized by the reference value. 
    A positive value indicates that $\hat{P}_F$ is an overestimate of the true value.
    A \textit{lower} absolute relative error indicates better performance.
    \item \textit{Average negative log likelihood (NLL)}: An evaluation batch of samples $\hat{\vec{x}}$ is drawn from the learned proposal distribution, and for each sample the negative log-likelihood is computed under the prior distribution. 
    A \textit{lower} average NLL indicates that the samples are more likely under the prior distribution.
    \item \textit{Coverage}: The coverage metric proposed by \citet{naeem2020reliable} measures the proportion of real samples whose neighborhood (as defined by its $k$-nearest neighbors) contains at least one generated sample.
    A \textit{higher} value indicates that more modes of the real samples are represented in the samples drawn from the learned proposal.
    \item \textit{Number of statistically different bins (NDB)}: \citet{richardson2018gans} propose the NDB metric as a simple method to evaluate generative models based on relative proportions of samples that fall into $C$ predetermined bins. A \textit{lower} value indicates that learned distribution more closely represents the data distribution.
    We report NDB/$C$ to normalize the values between $0$ and $1$.
    \item \textit{Total number of samples} ($N_{\text{total}}$): The total number of samples is a proxy for the sample efficiency of each method. \textit{Fewer} total samples indicate higher efficiency in finding an effective proposal distribution.
\end{itemize}

\subsection{Experimental Setup}
We evaluate MPPCA proposal densities against GMM proposals with full-rank covariance matrices.
The proposals are iteratively refined using the CE and SIS methods until the data log-likelihood converges.
We perform IS on the Branches problem with $40$ and $60$ dimensions, and on the Duffing Oscillator problem with $100$ and $200$ dimensions.
Problem hyperparameters are provided in \cref{tab:hyperparameters}.
All experiments run on a CPU, but parallelizing trials can improve efficiency.
Code to reproduce the experimental results is available at \texttt{\url{https://github.com/sisl/MPPCAImportanceSampling}}.

\subsection{Results}
\begin{table*}[t]
\caption{Results} 
\label{tab:metrics}
\centering
    \sisetup{round-mode=places, round-precision=4, table-align-uncertainty=true, separate-uncertainty=true}
    \begin{tabular*}{\textwidth}{@{\extracolsep{\fill}}
        c
        l
        S[table-format=2.3(4), separate-uncertainty=true, retain-zero-uncertainty = true]
        S[table-format=3.3(5), separate-uncertainty=true, retain-zero-uncertainty = true]
        S[table-format=1.3(4), separate-uncertainty=true, retain-zero-uncertainty = true]
        S[table-format=1.3(4), separate-uncertainty=true, retain-zero-uncertainty = true]
        S[table-format=6(5), separate-uncertainty=true, retain-zero-uncertainty = true]
    @{}}
    \toprule
     {Dataset} & {IS Method} & {$(\hat{P}_F - P_F)/P_F$} & {Avg NLL$(\hat{x})$} & {Coverage} & {NDB/$C$} & {$N_\text{total}$}\\
    \midrule
    \multirow{4}{*}{\shortstack[*]{\textbf{\branches{}}\\$d=40$\\$P_F = \num[round-precision=2]{9.55e-4}$}} 
        & CE-MPPCA  & \bfseries -0.024 \pm 0.018 & 65.660 \pm 0.191           & \bfseries 0.943 \pm 0.004 & \bfseries 0.128 \pm 0.049 & \bfseries 30000 \pm 0 \\
        & SIS-MPPCA & -0.043 \pm 0.096           & 66.976 \pm 0.365           & 0.760 \pm 0.015           & 0.707 \pm 0.059           & 83200 \pm 8352        \\
        & CE-GMM    & -0.997 \pm 0.006           & \bfseries 64.270 \pm 4.676 & 0.686 \pm 0.199           & 0.938 \pm 0.062           & 39200 \pm 2712        \\
        & SIS-GMM   & -0.053 \pm 0.593           & 73.533 \pm 2.223           & 0.667 \pm 0.037           & 0.731 \pm 0.060           & 84400 \pm 9200        \\
    \midrule
    \multirow{4}{*}{\shortstack[*]{\textbf{\branches{}}\\$d=$ 60\\$P_F = \num[round-precision=2]{9.33e-4}$}} 
        & CE-MPPCA  & \bfseries 0.001 \pm 0.027 & \bfseries 94.034 \pm 0.185 & \bfseries 0.944 \pm 0.005 & \bfseries 0.167 \pm 0.050 & \bfseries 30000 \pm 0 \\
        & SIS-MPPCA & -0.003 \pm 0.045          & 95.435 \pm 0.429           & 0.855 \pm 0.025           & 0.575 \pm 0.073           & 82800 \pm 6939        \\
        & CE-GMM    & -0.999 \pm 0.002          & 95.074 \pm 8.768           & 0.592 \pm 0.232           & 0.961 \pm 0.044           & 37600 \pm 4270        \\
        & SIS-GMM   & 0.204 \pm 2.100           & 103.779 \pm 1.513          & 0.709 \pm 0.033           & 0.716 \pm 0.080           & 82800 \pm 6939        \\
    \midrule
    \multirow{4}{*}{\shortstack[*]{\textbf{\duffing{}}\\$d=100$\\$P_F = \num[round-precision=2]{9.55e-4}$}} 
        & CE-MPPCA  & -0.012 \pm 0.024           & \bfseries 149.890 \pm 0.321 & \bfseries 0.945 \pm 0.007 & \bfseries 0.126 \pm 0.057 & \bfseries 30000 \pm 0 \\
        & SIS-MPPCA & \bfseries -0.008 \pm 0.050 & 152.536 \pm 0.784           & 0.937 \pm 0.011           & 0.297 \pm 0.120           & 127200 \pm 23919      \\
        & CE-GMM    & -0.999 \pm 0.000           & 165.675 \pm 17.590          & 0.918 \pm 0.122           & 0.894 \pm 0.069           & 38600 \pm 6003        \\
        & SIS-GMM   & -0.823 \pm 0.259           & 156.262 \pm 2.302           & 0.876 \pm 0.062           & 0.676 \pm 0.128           & 118000 \pm 20099      \\
    \midrule
    \multirow{4}{*}{\shortstack[*]{\textbf{\duffing{}}\\$d=200$\\$P_F = \num[round-precision=2]{9.39e-4}$}} 
        & CE-MPPCA  & 0.004 \pm 0.031            & \bfseries 291.721 \pm 0.664 & \bfseries 0.953 \pm 0.008 & \bfseries 0.201 \pm 0.062 & \bfseries 30000 \pm 0 \\
        & SIS-MPPCA & \bfseries -0.003 \pm 0.079 & 294.698 \pm 0.773           & 0.950 \pm 0.009           & 0.298 \pm 0.095           & 138800 \pm 30570      \\
        & CE-GMM    & -0.999 \pm 0.000           & 313.461 \pm 33.544          & 0.945 \pm 0.137           & 0.935 \pm 0.109           & 38000 \pm 9591        \\
        & SIS-GMM   & -0.999 \pm 0.001           & 303.757 \pm 3.464           & 0.745 \pm 0.120           & 0.785 \pm 0.104           & 123600 \pm 17292      \\
    \midrule
    \multirow{4}{*}{\shortstack[*]{\textbf{\gcas{}}\\$d=202$\\$P_F = \num[round-precision=2]{6.11e-4}$}}  
        & CE-MPPCA  & \bfseries -0.492 \pm 0.624 & 292.656 \pm 4.816           & 0.517 \pm 0.177           & \bfseries 0.936 \pm 0.044 & \bfseries 69000 \pm 11357 \\
        & SIS-MPPCA & -0.642 \pm 0.207           & 297.620 \pm 2.484           & 0.601 \pm 0.042           & 0.961 \pm 0.029           & 158800 \pm 26354          \\
        & CE-GMM    & -0.999 \pm 0.000           & 329.508 \pm 58.463          & 0.654 \pm 0.341           & 0.981 \pm 0.023           & 108600 \pm 76026          \\
        & SIS-GMM   & -0.998 \pm 0.000           & \bfseries 288.815 \pm 1.089 & \bfseries 0.893 \pm 0.021 & 0.975 \pm 0.021           & 171600 \pm 23009          \\
    \bottomrule
    \end{tabular*}
\end{table*}%

\Cref{tab:metrics} presents the experimental results across the five problem configurations. 
MPPCA proposals fit via the cross-entropy method (CE-MPPCA) obtain the smallest relative error of $\hat{P}_F$ on three of the five experimental configurations. 
MPPCA proposals fit via SIS obtain the smallest relative error on the remaining two problems.
This indicates that IS with MPPCA proposals provide more reliable probability estimates compared to GMM proposals.
Both CE-MPPCA and SIS-MPPCA score lower on the NDB/$C$ metric than CE-GMM and SIS-GMM, which indicates that samples from the learned MPPCA proposals are a closer match to samples from the true failure distributions.
The performance of different methods varies more for the average NLL metric, with no single approach consistently outperforming the others in all cases.
Note that both average NLL and coverage are computed in \textit{disturbance space} for the oscillator and F-16 problems, e.g., the proposals are over system inputs rather than output trajectories.
Even though the GMM proposals obtain higher average coverage scores on the F-16 problem, a visual inspection of \cref{fig:results} reveals that the GMMs experience a mode collapse.
The learned GMM disturbance models only induce failures due to aerodynamic stalls, while the MPPCA proposals learn disturbance distributions that induce both stalls and ground collisions.
The CE method converges with fewer samples than SIS, regardless of proposal density. 
This is likely because SIS relies on MCMC to gradually move samples towards the optimal failure distribution.

%% file: s6-conclusion.tex
\section{Conclusion and Future Work}
\label{sec:conclusion}
Importance sampling is a powerful technique for validating safety-critical systems \cite{corso2021survey}. 
In this work we present a technique to scale parametric IS methods using low-rank mixture proposal distributions.
Mixtures of probabilistic principal component analyzers are parameter-efficient models that can be updated analytically via expectation-maximization, even in high-dimensional spaces.
Closed-form parameter updates are critical for fast and efficient adaptive IS methods.
We empirically demonstrated that MPPCA proposal distributions obtain more reliable estimates of the probability of failure compared to GMM proposals on a range of challenging, high-dimensional systems.

Future work will explore connections between the number of system failure modes and the choice of latent MPPCA factors. 
Characterizing the principal subspace of the failure distribution will allow us to identify a meaningful number of latent factors for each probabilistic principal component analyzer model, refining the trade-off between model complexity and failure distribution fidelity.
We will also develop generative performance metrics that consider sample density and coverage in both disturbance space and trajectory space.